\documentclass[conference]{IEEEtran}
\usepackage{times}
\usepackage{latexsym}
\usepackage{booktabs}
\usepackage{helvet}
\usepackage{courier}

\usepackage{graphicx} 

\usepackage{amsmath}

\begin{document}
	%
	\title{Communication-Free Parallel Supervised Topic Models}
	\author{\IEEEauthorblockN{Lili(Lee) Gao}
		\IEEEauthorblockA{
			AQR Capital Management LLC\\
			2 Greenwich Plaza\\
			Greenwich, CT, 06830, USA\\
			lee.gao@aqr.com}
		\and\IEEEauthorblockN{Ronghuo Zheng}
		\IEEEauthorblockA{
			University of Texas at Austin\\
			2110 Speedway\\
			Austin, TX 78705, USA\\
			ronghuo.zheng@mccombs.utexas.edu}
	}
	\maketitle
\begin{abstract}
Embarrassingly (communication-free) parallel Markov chain Monte
Carlo (MCMC) methods are commonly used in learning graphical models. However, MCMC cannot be directly applied in learning topic models because of the quasi-ergodicity problem caused by multimodal distribution of topics. 
In this paper, we develop an embarrassingly parallel MCMC algorithm for sLDA. Our algorithm works by switching the order of sampled topics combination and labeling variable prediction in sLDA, in which it overcomes the quasi-ergodicity problem because high-dimension topics that follow a multimodal distribution are projected into one-dimension document labels that follow a unimodal distribution.  
Our empirical experiments confirm that the out-of-sample prediction performance using our embarrassingly parallel algorithm is comparable to non-parallel sLDA while the computation time is significantly reduced.
\end{abstract}

\section{Introduction}

Topic modelling is one of the most important and widely used machine learning tools. It
has been successfully applied to various types of problems, including text analysis, image processing, speech recognition, and more. Topic models are useful in learning the hierarchical structures of data. For example, when applied to text documents, latent Dirichlet allocation (LDA)~\cite{blei2003latent} is a generative
probabilistic topic model for collections of text corpora. In this application, each
document is modeled as a finite mixture over a set of underlying topics,
and each topic is modeled as a distribution over words. 

Supervised latent Dirichlet allocation (sLDA)~\cite{mcauliffe2008supervised}
is used to infer latent topics predictive of document labeling variables.
For example, sLDA has been used to predict star ratings using movie reviews, or to predict e-commerce sales using customer reviews, and so on. Along with the exponential growth of data, especially the text data and images on the Web, sLDA is often used in applications dealing with big data. The large-scale learning of sLDA is generally slow and
difficult. A natural way to speed up the computation is to
resort to parallel computing. In this paper, we aim at designing an efficient parallel algorithm for large-scale learning of sLDA.

Communication-free parallel algorithm are commonly used in machine learning, which minimizes communication costs resulted from learning
synchronization. 
Recent
research~\cite{neiswanger2013asymptotically,wang2013parallel,minsker2014scalable, xu2014distributed, calderhead2014general}
proposed a parallel Markov chain Monte Carlo (MCMC) algorithm with
little communication cost, and showed that it can significantly improve the computation speed. The basic
idea is to independently run a separate MCMC sampler to generate a local
sample set for each sub-dataset, and combine the local sample sets to estimate the desired global posterior distribution. However, existing algorithms have limitations in learning graphical models with multimodal posteriors because of the quasi-ergodicity problem. It means that although theoretically a sampler has positive probability of switching from a local mode to another, the sampler may get stuck in a local mode numerically. If the local posteriors converge to different local modes and differ significantly, the global posterior representation
achieved in the final combination stage can be highly inaccurate. The posterior of topics in (s)LDA is multimodal because each permutation of the topics forms a new mode. Hence, the existing parallel technique
cannot be directly applied to sLDA.

In this paper, we develop a parallel algorithm for sLDA which maintains the speed advantage of embarrassingly parallel computing while
at the same time bypassing the problematic quasi-ergodicity problem. The idea is that since the main objective of sLDA is prediction rather than mere categorization, it is not necessary to estimate the global posterior of document topics. It allows us to switch the order of document topics combination and labeling variable prediction and thus overcome the quasi-ergodicity problem cause by multimodal posterior of sLDA. Specifically, instead of firstly combining the sub sampled topics generated by the parallel samplers to form a sample of the global posterior distribution and secondly making predictions based on the estimated global posterior, we first make predictions using sub-posteriors based on each
sub-dataset and then combine the local predictions to make the global prediction. By making prediction first, we can project the high-dimensional topics that follow a multimodal distribution into the one-dimensional document label that follow a unimodal distribution. Hence, we overcome the quasi-ergodicity problem in the combination stage.

To implement our parallel algorithm for sLDA, we propose two different combination rules to combine the local predictions to the global predictions, i.e., \emph{Simple Average} and \emph{Weighted Average}.
In Simple Average, we first apply MCMC in parallel to sample topics for the words and make predictions of the test dataset using each of
the locally estimated sLDA model, called local predictions. After that, 
we take the simple arithmetic average of the local predictions as
our global predictions (i.e., predictions using the whole training
dataset). In Weighted Average, the only difference from Simple Average
is the way we combine local predictions into global predictions. Instead
of taking simple arithmetic average, we take a weighted average of
the local predictions. The weight for the local predictions
is equal to either the inverse of the corresponding training dataset mean square errors (MSEs) or the prediction accuracy of the corresponding training dataset.

We conduct two empirical experiments to evaluate our parallel algorithm for sLDA. In first experiment we use textual documents of corporate annual reports to predict the labeling variable earnings per share. In the second experiment we use the textual documents of IMDB movie reviews to predict the labeling variable, sentiment scores based on the IMDB ratings. The empirical performance confirms that the test
set mean square error and prediction accuracy using our embarrassingly parallel algorithm is comparable to a single machine sLDA while the computation time is significantly reduced.

The rest of the paper is organized as follows. Section~\ref{sec:related_work} reviews the recent literatures on parallel MCMC and sLDA. Section~\ref{sec:epMCMC} introduces our embarrassingly parallel MCMC method for sLDA. Section~\ref{sec:empirical_results} shows the empirical results and Section~\ref{sec:conclusion} concludes.

%

\section{Related Work}\label{sec:related_work}

The paper is related to recent literatures on parallel MCMC.
\cite{neiswanger2013asymptotically} presented a parallel MCMC algorithm
that requires little communication but still maintains asymptotic
correctness. The idea is to first partition the dataset into multiple
sub-datasets, and then perform MCMC sampling in parallel using only the sub-dataset
on each machine, and finally combine the sub-samples to construct
the full samples. In \cite{neiswanger2013asymptotically}, Gaussian
models are assumed for each local posterior, which allows for an explicit
product of densities in the final combination stage. Using a similar idea, \cite{wang2013parallel}
makes one step further by representing the posterior as a Weierstrass transform.
They analytically show that the approximation error for the Weierstrass
sampler is bounded by some tuning parameters and provide suggestions
for choice of the values. The simulation result confirms its computation
efficiency. \cite{minsker2014scalable} improve the final combination stage in
the presence of outliers by using the median posterior (i.e., the geometric
median of subset posterior distributions). Specifically, \cite{minsker2014scalable}
proposed a Bayesian inference approach that is scalable and robust
to corruption in the data, which allows to avoid extensive communication among machines by running independent
MCMC chains for split subsets. The novelty is that they proposed
a different approximation to the full data posterior for each subset
based on the evaluation of the geometric median of subset posterior
distributions. Our paper also focuses on the posterior combination stage in order to overcome the quasi-ergodicity problem.

This paper is also related to another stream of literature on (s)LDA.
Since being introduced in \cite{blei2003latent}, LDA has been widely used in text mining and machine learning.  In~\cite{mcauliffe2008supervised}, supervised latent Dirichlet
allocation (sLDA) is introduced to infer latent topics predictive
of the response. Compared with LDA in
\cite{blei2003latent}, sLDA is more appropriate for prediction rather
than mere categorization. For parameter estimation, an approximated maximum-likelihood
estimation is carried out using a variational Expectation-Maximization
(EM) procedure to handle intractable posterior expectations. The sLDA
model proposed in~\cite{mcauliffe2008supervised} can accommodate various types of response. However, the parameter estimation using Expectation-Maximization (EM) algorithm
is susceptible to problems involving local maxima and slow convergence
rate. \cite{griffiths2004finding} develop a MCMC method for LDA using
collapsed Gibbs sampling. The method has been widely used and has also
been adapted for learning sLDA~\cite{nguyen2014sometimes}.
In order to increase computational efficiency of learning large-scale sLDA, a natural idea is
to resort to parallel techniques.
However, as the posteriors in LDA are multimodal,
parallel sampling can be problematic because of the quasi-ergodicity problem. In this paper, we overcome the quasi-ergodicity
problem for embarrassingly parallel sLDA by focusing on prediction rather than
topic categorization.

\section{Embarrassingly Parallel MCMC for sLDA}\label{sec:epMCMC}

We start with introducing the embarrassingly parallel MCMC method and illustrating
the quasi-ergodicity problem for learning multimodal graphical models, such as sLDA. Then we introduce
the single machine nonparallel MCMC method for supervised topic models, on which we base our paralleled MCMC method.
After that, we propose
our embarrassingly paralleled MCMC method for sLDA which overcomes the quasi-ergodicity problem.

\subsection{Quasi-ergodicity of Embarrassingly Parallel MCMC}

\begin{figure}[ht]
\centering 
 \includegraphics[width=0.45\textwidth]{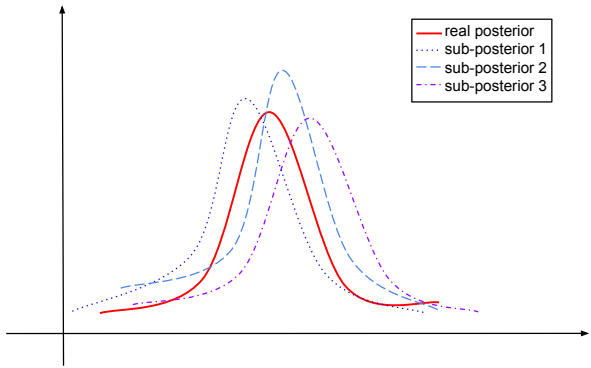}\\
 \protect\caption{The Effectiveness of Embarrassingly Parallel MCMC for Unimodal}

\label{Fig_singlemodal}
\end{figure}

Recent research (\cite{neiswanger2013asymptotically,wang2013parallel,minsker2014scalable})
has shown that for models with certain properties,
embarrassingly parallel MCMC can guarantee asymptotic property theoretically and perform well empirically.
The basic idea of embarrassingly parallel MCMC is to independently run a Gibbs
sampler in parallel for each sub-dataset
and then combine the local sub-samples to global posterior distribution 
(see illustration in Figure~\ref{Fig_singlemodal}). Following this basic idea, a naive application of embarrassingly parallel MCMC on sLDA includes the following
four steps:\footnote{For convenience in algorithm descriptions, we assume we apply sLDA to model
textual topics for text documents, which is the most popular
application of sLDA.}
\begin{enumerate}
\item Divide the training dataset into several subsets.
\item Run MCMC in parallel to sample textual topics for the documents in each subset.
\item Combine the sampled topics for each subset and treat the combined samples
as if they were directly sampled using all documents in the whole dataset.
\item Estimate sLDA parameters using the combined topics and
make predictions based on the estimated model.
\end{enumerate}
However, in the presence of multimodal posteriors, which is often true for models
with a large number of latent variables such as LDA (because there
exists a mode for each permutation of the topic labels), embarrassingly
parallel MCMC can fail due to quasi-ergodicity. As a result, the global posterior
representation achieved in the sample combination stage can be highly
inaccurate.

\begin{figure}[ht]
\centering 
 \includegraphics[width=0.45\textwidth]{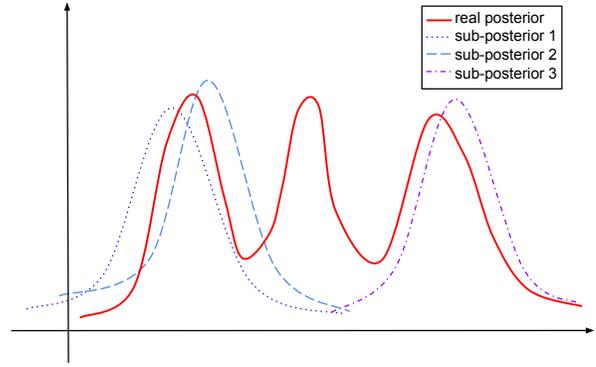}\\
 \protect\caption{The Quasi-Ergodicity of Embarrassingly Parallel MCMC for Multimodal}

\label{Fig_multimodal}
\end{figure}

The intuition of the quasi-ergodicity problem can be shown in a comparison
of unimodal posteriors illustrated in Figure \ref{Fig_singlemodal} and multimodal posteriors illustrated in Figure \ref{Fig_multimodal}. 
In Figure \ref{Fig_singlemodal}, the true posterior is a unimodal
distribution. The parallel computing uses three machines, and each machine generates some sub-samples whose empirical
distribution is close to the
true posterior. Therefore, the empirical distribution of the combined samples
constitutes a valid estimation of the true
posterior. In Figure \ref{Fig_multimodal}, the true posterior is
a three-modal distribution.   The parallel computing also uses three machines. With multimodal posteriors, it is very likely that samples generated by different
machines will converge to different modes, especially in high-dimensional space. In the example shown in Figure \ref{Fig_multimodal}, two machines turn out to sample from the leftmost
mode and the third machine samples from the rightmost mode. For a single machine, converging to any one
of the three modes catches one peak of the true posterior,
and thus can be considered as a consistent posterior.
However, it is problematic to naively combine the posterior samples generated by
three machines converging to different modes. In this case, the
combined posterior might not have significant peak but have a wide flat region
which covers both peak and nadir in the true posterior. The resulting
empirical posterior would be significantly different from the true posterior.

The quasi-ergodicity problem seems difficult to avoid for LDA in which
the objective is to learn the categorization of topics. However, the
quasi-ergodicity problem can be bypassed in sLDA in which the main objective
is to make predictions based on topics rather than learning the topic categorization.

\begin{figure}[ht]
\centering 
 \includegraphics[width=0.45\textwidth]{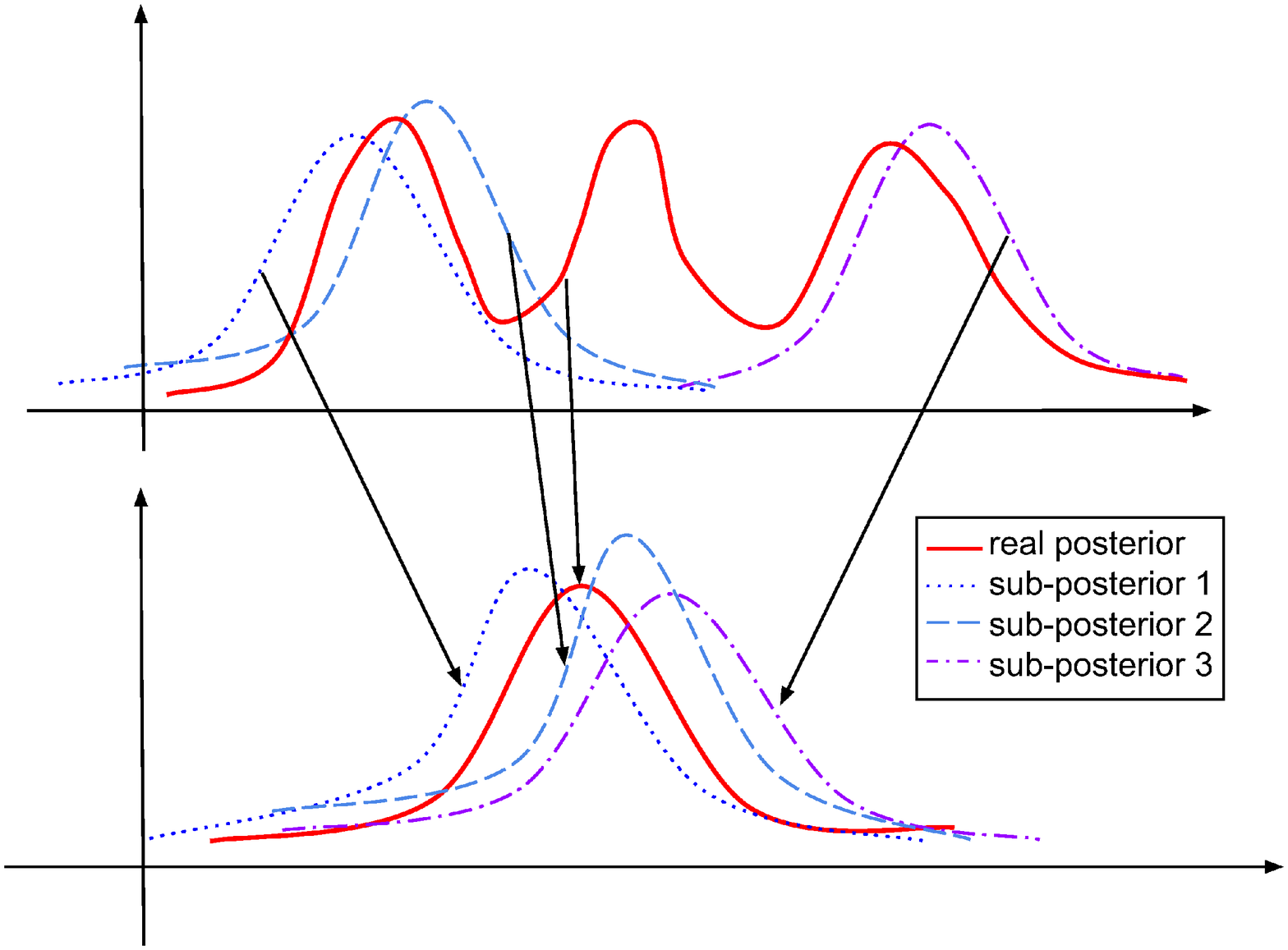}\\
 \protect\caption{Solving Quasi-Ergodicity of Embarrassingly Parallel MCMC for sLDA}

\label{Fig_multi2single}
\end{figure}

Our proposed solution is to switch the order of sub-sample combination
and prediction. In particular, instead of first combining sub-samples generated by running
MCMC on each sub-dataset to constitute the samples
for full training dataset and then using the combined sample to make predictions of the document label, we first make predictions based on sub-samples and then combine these predictions using our proposed combination rules: Simple Average and Weighted
Average. The intuition is
illustrated in Figure \ref{Fig_multi2single} about why switching the order of subset combination
and prediction can help to bypass the quasi-ergodicity problem. By making predictions first, we
can project the high-dimensional topics that follow a multimodal distribution into a one-dimensional document labels that follow a unimodal distribution.

\subsection{MCMC for sLDA}

\label{sec_upMCMC}

Before we describe our embarrassingly parallel MCMC algorithm for sLDA, we first review the basic notations, the background,
and the existing single machine non-parallel MCMC method of sLDA.

\begin{figure}[ht]
\centering 
 \includegraphics[width=0.45\textwidth]{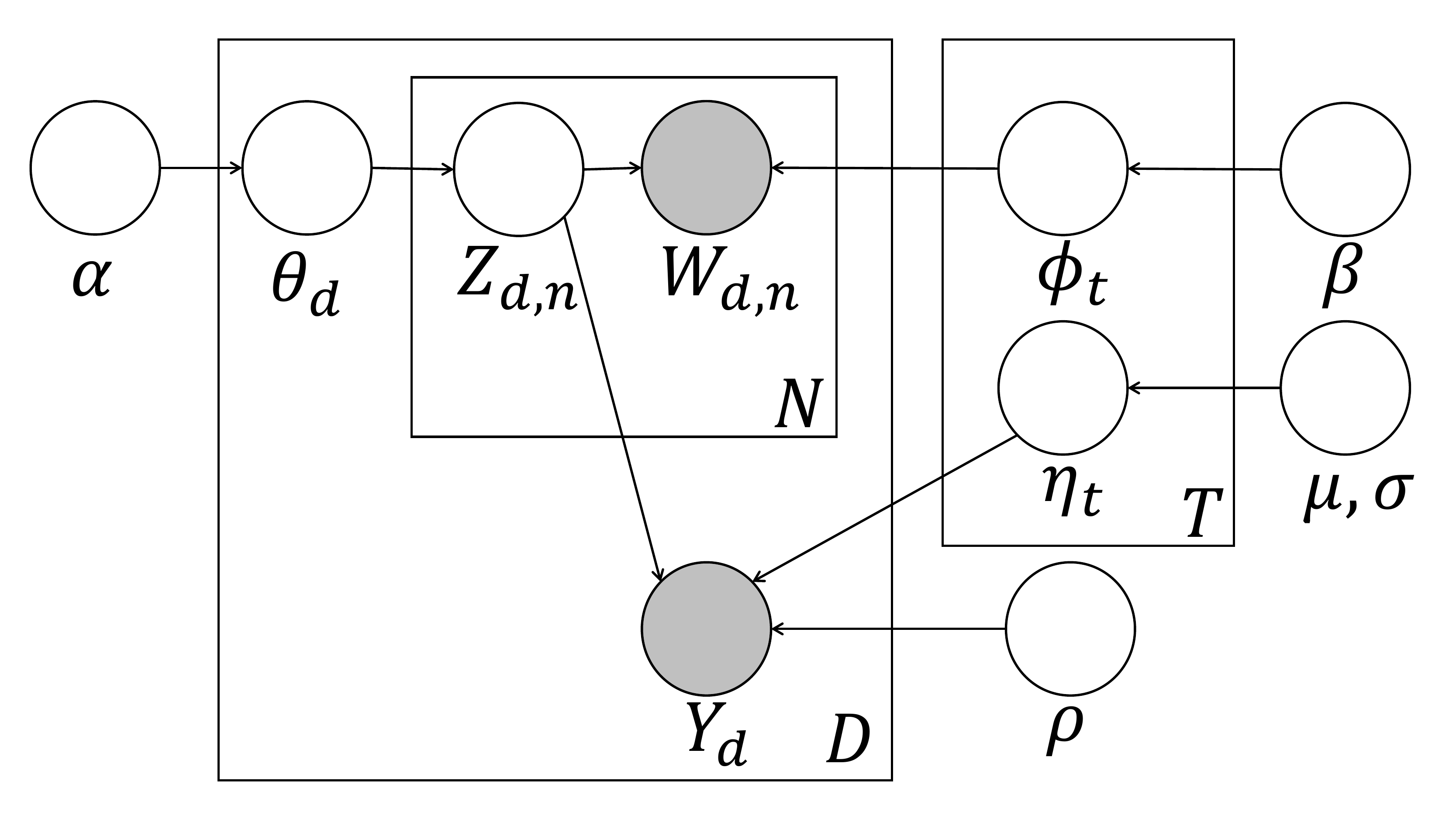}\\
 \protect\caption{Graphical Model of sLDA}

\label{Fig_slda}
\end{figure}

The generative process of sLDA~\cite{mcauliffe2008supervised} is as follows (the graphical structure is shown in Fig~\ref{Fig_slda}):
\begin{enumerate}
\item For each topic $t\in\{1,2,\cdots,T\}$:

\begin{enumerate}
\item Draw word distribution $\phi_{t}\sim Dir(\beta)$;
\item Draw parameter $\eta_{t}\sim\mathcal{N}(\mu,\sigma)$.
\end{enumerate}
\item For each document $d\in\{1,2,\cdots,D\}$:

\begin{enumerate}
\item Draw topic distribution $\theta_{d}\sim Dir(\alpha)$;
\item For each word $n\in\{1,2,\cdots,N_{d}\}$:

\begin{enumerate}
\item Draw topic $z_{d,n}\sim Multi(\theta_{d})$;
\item Draw word $w_{d,n}\sim Multi(\phi_{z_{d,n}})$.
\end{enumerate}
\item Draw response $y_{d}\sim\mathcal{N}(\mathbf{\eta^{\top}\bar{z}_{d}},\rho)$
where $\bar{z}_{d,t}=\frac{1}{N_{d}}\sum_{n=1}^{N_{d}}\mathbf{1}_{t}(z_{d,n})$
and $\mathbf{1}_{t}(z_{d,n})$ is an indicator function, i.e., $\mathbf{1}_{t}(z_{d,n})=1$
if $z_{d,n}=t$ and $0$ otherwise.
\end{enumerate}
\end{enumerate}
The Gibbs sampling method for LDA was first introduced by~\cite{griffiths2004finding},
and~\cite{nguyen2014sometimes} modified the method for sLDA. In
this paper, we follow~\cite{nguyen2014sometimes} for both posterior
inference and prediction using MCMC in supervised topic modeling.
The data is divided into training data (TR) for posterior inference
and testing data (TE) for prediction.

\subsubsection{Posterior Inference}

There are two alternative steps for posterior inference using stochastic
EM. The first step is a Gibbs sampling to assign a topic $z_{d,n}$
to each token $w_{d,n}$. The second step is to optimize the regression
parameters $\eta$.

\paragraph{Gibbs Sampling}

The probability of assigning topic $t$ to
token $w_{d,n}^{\text{TR}}=w$ in training document $d$ is:
\begin{align}
p\left(z_{d,n}^{\text{TR}}=t|\mathbf{z}_{-d,n}^{\text{TR}},\mathbf{w}^{\text{TR}},y_{d}\right) & \propto\nonumber \\
N\left(y_{d};\mu_{d,n},\rho\right)\cdot
\frac{N_{\text{TR},d,t}^{-d,n}+\alpha}{N_{\text{TR},d,\cdot}^{-d,n}+T\alpha} & \cdot\frac{N_{\text{TR},t,w}^{-d,n}+\beta}{N_{\text{TR},t,\cdot}^{-d,n}+W\beta},
\end{align}
where $\mu_{d,n}=\left(\sum_{t'=1}^{T}\eta_{t'}N_{\text{TR},d,t'}^{-d,n}+\eta_{k}\right)/N_{\text{TR},d}$
is the mean of the Gaussian generating $y_{d}$ if $z_{d,n}^{\text{TR}}=t$
and $N_{\text{TR},\cdot,\cdot}^{-d,n}$ is a count that does not include
the current assignment of $z_{d,n}^{\text{TR}}$. The first term, $N\left(y_{d};\mu_{d,n},\rho\right)$, is
the margin of the response; the second term, $\frac{N_{\text{TR},d,t}^{-d,n}+\alpha}{N_{\text{TR},d,\cdot}^{-d,n}+T\alpha}$, is the ratio representing
the probability of topic $t$ in document $d$; and the third term, $\frac{N_{\text{TR},t,w}^{-d,n}+\beta}{N_{\text{TR},t,\cdot}^{-d,n}+W\beta}$, 
is the ratio representing the probability of word $w$ under topic
$t$.

\paragraph{Optimizing the Regression Parameters}

The parameters $\mathbf{\eta}$
is estimated by maximizing the likelihood:
\begin{small}
\begin{align}
L\left(\mathbf{\eta}\right) & =-\frac{1}{2\rho}\sum_{d=1}^{D}\left(y_{d}^{\text{TR}}-\mathbf{\eta}^{\top}\bar{\mathbf{z}}_{d}^{\text{TR}}\right)^{2}
-\frac{1}{2\sigma}\sum_{t=1}^{T}\left(\eta_{t}-\mu\right)^{2}.
\end{align}
\end{small}

The alternative iteration converges with the estimated parameters
$\hat{\mathbf{\eta}}$, and the distribution over words $\hat{\phi}_{t}$
of topic $t$ is estimated as:
\begin{align}
\hat{\phi}_{t,w} & =\frac{N_{\text{TR},t,w}+\beta}{N_{\text{TR},t,\cdot}+W\beta}.
\end{align}

\subsubsection{Prediction}

For prediction for the test data, we first sample the topic assignments
for all tokens in the test data:
\begin{align}
p\left(z_{d,n}^{\text{TE}}=t|\mathbf{z}_{-d,n}^{\text{TE}},\mathbf{w}^{\text{TE}}\right) & \propto\frac{N_{\text{TE},d,t}^{-d,n}+\alpha}{N_{\text{TE},d,\cdot}^{-d,n}+T\alpha}\hat{\phi}_{t,w_{d,n}^{\text{TE}}}
\end{align}
 The predicted value is the estimated value of response:
\begin{align}
\hat{y}_{d}^{\text{TE}} & =\mathbf{\hat{\mathbf{\eta}}}^{\top}\bar{\mathbf{z}}_{d}^{\text{TE}}
\end{align}
 where $\bar{\mathbf{z}}_{d,t}^{\text{TE}}\equiv\frac{1}{N_{\text{TE},d}}\sum_{n=1}^{N_{\text{TE},d}}\mathbf{1}_{t}\left(z_{d,n}^{\text{TE}}\right)$
is the empirical topic distribution of test document $d$.

Note that when the label $y_D$ is discrete, we can model the logit probability of the label $y_D$ to be normally distributed. The corresponding MCMC method for sLDA can be achieved by slightly adjusting the procedures described above.

\subsection{Embarrassingly Parallel MCMC for sLDA}

After illustrating the procedure of unparalleled MCMC algorithm of sLDA,
we move to the procedure of embarrassingly
parallel MCMC algorithm of sLDA. As a benchmark, we first introduce the Naive Combination
which intuitively and naively follows the embarrassingly parallel MCMC by combining sub-posterior rather than
sub-predictions.

\paragraph{\textbf{Naive Combination}}
\begin{enumerate}
\item Divide the documents to training documents $D^{\text{TR}}$ and testing
documents $D^{\text{TE}}$. Partition the training documents into
$M$ subset $\{D^{\text{TR},(1)},D^{\text{TR},(2)},\cdots,D^{\text{TR},(M)}\}$;
\item For $m=1,2,\cdots,M$ (in parallel): assign each token a topic through Gibbs sampling;
\item Conduct posterior inference by treating the combination of sub-sample topics
as if they were directly sampled for the whole training sample:

\begin{enumerate}
\item obtain the estimated parameters $\hat{\mathbf{\eta}}$ by ordinary
linear regression;
\item the distribution over words $\hat{\phi}_{t}$ of topic $t$;
\end{enumerate}
\item Prediction: Gibbs sampling using $\hat{\phi}_{t}$ to obtain the empirical
topic distribution of test document $d$ as $\bar{\mathbf{z}}_{d}^{\text{TE}}$,
and estimate the prediction for document $d$ as $\hat{y}_{d}^{\text{TE}}=\left(\mathbf{\hat{\mathbf{\eta}}}\right)^{\top}\bar{\mathbf{z}}_{d}^{\text{TE}}$.
\end{enumerate}

As discussed above, the Naive Combination has the quasi-ergodicity problem because of the multimodal posterior of LDA.
We propose our embarrassingly parallel MCMC for sLDA to overcome the quasi-ergodicity problem
by leveraging the fact that the main
objective of sLDA is prediction rather than mere categorization. For prediction
in sLDA, it is not necessary to generate a global posterior about
the topics. Our method is to first run prediction using the sub-posterior based
on each sub-dataset and then combine the local predictions to the global
predictions, instead of first combining
the sub-posterior to a global posterior and then run prediction using the
global posterior as suggested in the Naive Combination. The formal procedure of our embarrassingly parallel algorithm
is described as follows:

\paragraph{\textbf{Embarrassingly parallel for sLDA}}
\begin{enumerate}
\item Divide the documents to training documents $D^{\text{TR}}$ and testing
documents $D^{\text{TE}}$. Partition the training documents into
$M$ subset $\{D^{\text{TR},(1)},D^{\text{TR},(2)},\cdots,D^{\text{TR},(M)}\}$;
\item For $m=1,2,\cdots,M$ (in parallel):

\begin{enumerate}
\item Posterior inference: obtain the estimated parameters $\hat{\mathbf{\eta}}^{(m)}$
and the distribution over words $\hat{\phi}_{t}^{(m)}$ of topic $t$;
\item Prediction: Gibbs sampling using $\hat{\phi}_{t}^{(m)}$ to obtain
the empirical topic distribution of test document $d$ as $\bar{\mathbf{z}}_{d}^{\text{TE},(m)}$,
and estimate sub-prediction for document $d$ as $\hat{y}_{d}^{\text{TE},(m)}=\left(\mathbf{\hat{\mathbf{\eta}}}^{(m)}\right)^{\top}\bar{\mathbf{z}}_{d}^{\text{TE},(m)}$;
\end{enumerate}
\item Combine the sub-predictions to full-data prediction:
\begin{align}
\hat{y}_{d}^{\text{TE}}=\mathrm{Comb}(\hat{y}_{d}^{\text{TE},(1)},\hat{y}_{d}^{\text{TE},(2)},\cdots,\hat{y}_{d}^{\text{TE},(M)}).
\end{align}

\end{enumerate}
Now, we specify how to combine the local
predictions to the global predictions. We propose two combination
methods: Simple Average and Weighted Average.
The first two steps in the embarrassingly parallel procedure are the same for Simple
Average and Weighted Average. The two methods are different only in the
third step: they use different combination functions.

\paragraph{Simple Average}

In Simple Average, after making separate predictions for the test data set using topics sampled by parallel MCMC based on each sub-training data set,
we take the arithmetic average of the
local predictions to get the global predictions:
\begin{align}
\mathrm{Comb}(\hat{y}^{(1)},\hat{y}^{(2)},\cdots,\hat{y}^{(M)})=\frac{1}{M}\sum_{m=1}^{M}\hat{y}^{(m)}.
\end{align}

\paragraph{Weighted Average}

In Weighted Average, after making predictions based on topics sampled by parallel
MCMC based on each subset, we take the weighted
average of the local predictions to get
the global predictions. For example, the weight $w^{\left(m\right)}$ assigned
to each set of local predictions can be equal to the inverse of training
set Mean Square Error (MSE) for the sLDA learned based on the corresponding subset:

\begin{align}
w^{\left(m\right)} &= \frac{1/MSE^{\left(m\right)}}{\sum_{n=1}^{M}1/MSE^{\left(n\right)}}\\
\mathrm{Comb}(\hat{y}^{(1)},\hat{y}^{(2)},\cdots,\hat{y}^{(M)})&=\sum_{m=1}^{M}w^{\left(m\right)}\hat{y}^{(m)}.
\end{align}

The
training set MSE is generated by using the sLDA learned on each subset
to predict the dependent labels of the whole training set.

%

\section{Empirical Results}\label{sec:empirical_results}

In this section, we evaluate our algorithms by comparing our embarrassingly
parallel MCMC algorithms of sLDA (both Simple Average and Weighted Average)
with two benchmarks: (1) Non-parallel MCMC algorithm of sLDA, and
(2) Embarrassingly parallel MCMC algorithm of sLDA combining sub-posterior
rather than sub-predictions (Naive Combination). We include Non-parallel MCMC algorithm of sLDA as a benchmark to compare both the prediction accuracy and the computation time of our embarrassingly
parallel MCMC algorithms of sLDA. We include Naive Combination as another benchmark to empirically show that
naively
combining the sub-samples does not constitute a valid sample from
the posterior based on the whole sample in topic models and thus results in low prediction accuracy.

We conduct two groups of experiments using different data sets. The first group of experiments uses 
the documents in management discussion
and analysis (MD\&A) in 10-K report to predict the earnings per share in the
corresponding firm-year. The second group of experiments uses the text in movie reviews to predict the sentiment of the movie reviews.

\subsection{Description of Data Sets}
We first introduce the data sets used in our experiments.
\subsubsection{Experiment I - Management Discussion and Analysis}
In the United States, the Securities and Exchange Commission mandates
all public firms to file yearly corporate reports known as ``Form
10-K''. This report typically includes corporate organization and
history information, consolidated statements, as well as management
discussion and analysis (MD\&A), and our analysis focus on the text
information in the MD\&A section, where firm managers often express
their subjective opinions of a firm's operational performance.
The textual information is believed to be highly correlated
with variables that measure firm performance and has been investigated in lots of
work in financial economics (e.g., ~\cite{kogan2009predicting, loughran2011liability}).
The document labeling variable we use is earnings per share (EPS). Therefore,
our goal is to predict EPS based on management discussions using a sLDA model.

The 10-K reports are publicly available on the SEC Edgar database,
and earnings per share information is from Compustat database. We
use Central Index Key (CIK) as an identifier to match firms in both
databases.

We first downloaded all the 10-K reports in year 2012, extracted
the MD\&A section, tokenized the text, tagged the tokens using Stanford
Log-linear Part-Of-Speech Tagger \cite{toutanova2003feature}, and
generated adjective-noun phrases based on the tagged-tokens. As most
of the phrases appear very infrequently, we only included phrases
that appear in at least 2\% of the total number of firms. Our finalized
dataset contains 4216 firms and 4238 phrases.

\begin{figure}[ht]
\begin{centering}
\includegraphics[width=0.45\textwidth]{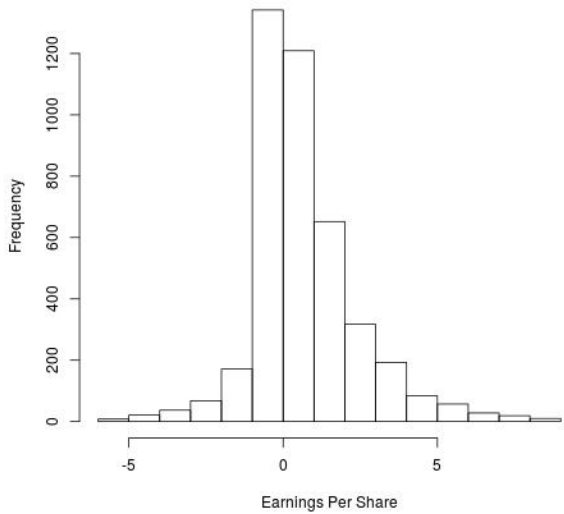}
\par\end{centering}

\protect\caption{Histogram of Earnings Per Share\label{fig:eps}}
\end{figure}

The histogram of earnings per share is shown in Figure \ref{fig:eps}.
We can see that the distribution of earnings per share is close to
normal distribution, implying it satisfy the normal assumption of
document label variable in the assumption for sLDA in \cite{mcauliffe2008supervised}.

\subsubsection{Experiment II - Movie Reviews} The data set consists of 50,000 IMDB movie reviews with labels.\footnote{The data set can be downloaded at https://www.kaggle.com/c/word2vec-nlp-tutorial/data.} Those movie reviews are specially selected for sentiment analysis. The label is the sentiment of the movie review, which is represented by a binary sentiment score. The sentiment score is $0$ if the IMDB rating is smaller than $5$ (i.e., negative reviews) and is $1$ if the IMDB rating is higher than $7$ (i.e., positive reviews). Our goal is to predict the sentiment score based on the text of movie reviews using a sLDA model.

The data set is divided into the 25,000 review labeled training set and the 25,000 review test set. No individual movie has more than 30 reviews.

\subsection{Evaluation}
Now we explain how we evaluate our algorithm in those two groups of experiments.
\subsubsection{Experiment I - Management Discussion and Analysis}
We randomly draw 3000 of the 4216 observations as the training set,
and the rest 1216 observations as the test set. We test the parallel
algorithms on a dual-core CPU computer with 4 threads, and thus for
each run of parallel computing, we randomly divide our training set
into 4 subset, 750 observations each.

\begin{figure}[h]
\begin{centering}
\includegraphics[width=0.45\textwidth]{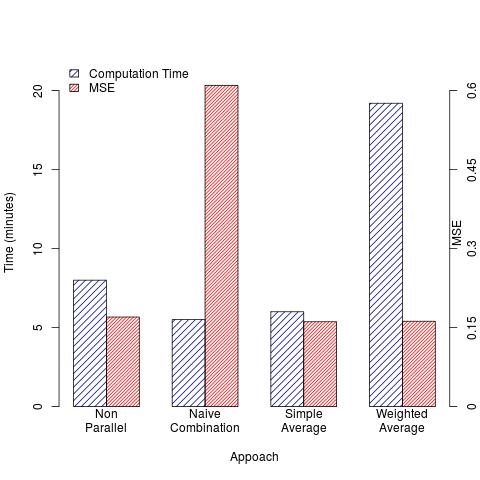}
\par\end{centering}

\protect\caption{Computational time and prediction accuracy comparison using MD\&A to predict earnings per share\label{fig:mse}}
\end{figure}

We evaluate the computational speed and prediction accuracy of the
four algorithms we mentioned above: Non-parallel, Naive Combination,
Simple Average and Weighted Average, by comparing the computational
time and test set prediction mean square error (MSE). The average results of 100 runs are shown in Figure \ref{fig:mse}.

\subsubsection{Experiment II - Movie Reviews} We only use the labeled training set containing 25,000 reviews in our experiment because there is no label for the test set to verify the accuracy. We randomly draw 20,000 observations (out of the 25,000 review labeled training set) as the training set and use it to predict the sentiment score of the rest 5,000 observations as the test set. We test the parallel
algorithms on a dual-core CPU computer with 4 threads, and thus for
each run of parallel computing, we randomly divide our training set
into 4 subset, 5,000 observations each.

\begin{figure}[h]
\begin{centering}
\includegraphics[width=0.45\textwidth]{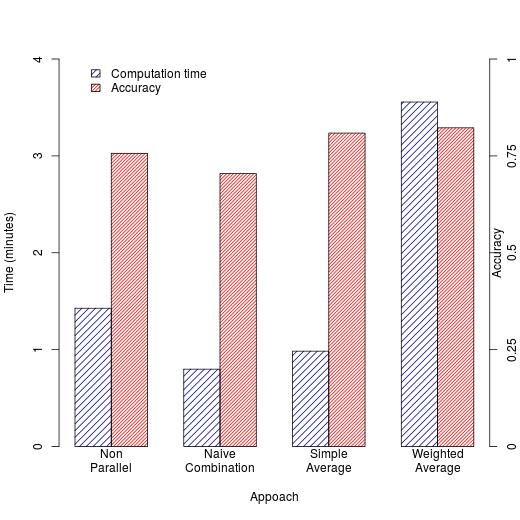}
\par\end{centering}

\protect\caption{Computational time and prediction accuracy comparison using movie reviews to predict sentiment score\label{fig:accu}}
\end{figure}

We evaluate the computational speed and prediction accuracy of the
four algorithms we mentioned above: Non-parallel, Naive Combination,
Simple Average and Weighted Average, by comparing the computational
time and test set prediction accuracy. The average results of 100 runs are
shown in Figure \ref{fig:accu}.

\subsubsection{Results and Discussion}
As shown
in Figure \ref{fig:mse} and Figure~\ref{fig:accu}, Naive Combination and Simple
Average are faster than the Non-parallel approach, but Weighted Average
is much slower than the other three. Naive Combination is the fastest
one because of the computational advantage of communication-free parallel
MCMC. Simple Average and Weighted Average also benefit from
the computational advantage of parallel MCMC. Whereas, in these two algorithms,
we need to make predictions for the test dataset based on the local
posteriors using each sub training dataset, which means four predictions
based on four different sub training dataset
are generated for each test data, while only one prediction is
necessary to be made for Naive Combination. The additional predictions
required by Simple Average and Weighted Average weaken the speed
advantage of parallel computing and, thus, Simple Average and Weighted Average are slower
than Naive Combination. Compared with Non-parallel, Simple Average is still much faster.
A drawback of Weighted Average is that it also needs to make predictions for the
training dataset in order to derive the weights, and the step is so costly when the
training dataset is large
that it is even slower than Non-parallel in our experiment.

Figure \ref{fig:mse} shows that the test dataset MSE of
Naive Combination is much larger than the other three algorithms. Similarly, Figure~\ref{fig:accu} shows that the test dataset prediction accuracy of Naive Combination is lower than the other three algorithms. These results
confirm the existence of quasi-ergodicity problem of LDA
models. Figure~\ref{fig:mse} also shows that the test dataset MSE of Simple Average and Weighted Average is
close to the Non-parallel benchmark. Figure~\ref{fig:accu} shows that the test dataset prediction accuracy of Simple Average and Weighted Average can be slightly higher than the Non-parallel benchmark. These results imply the quasi-ergodicity problem
is overcome by our proposed algorithms. Therefore, the proposed parallel
algorithms can indeed bypass the quasi-ergodicity problem for sLDA.

After all, when considering both the computation time
and prediction accuracy, Simple Average seems to be the most efficient approach. It speeds
up computation and also retains prediction accuracy. One computational
drawback of Simple Average is that we need to predict test dataset
in each parallel machine. This implies Simple Average can save more time compared with Non-parallel when we have large training
dataset and small test dataset.

\section{Conclusion}\label{sec:conclusion}

In this paper, we develop an embarrassingly parallel MCMC method for sLDA to bypass the quasi-ergodicity problem. A naive application of embarrassingly parallel MCMC for (s)LDA suffers from the quasi-ergodicity problem so a simple combination of sampled topics generated by parallel samplers for each sub training dataset does not constitute valid samples for the global posterior distribution. The quasi-ergodicity problem can be bypassed by switching the order of subset combination and prediction, meaning firstly making predictions using local sub-samples and then combining the local predictions to form the global prediction. This order switching is allowed in sLDA because the main objective is prediction rather than topic categorization. 

Based on our framework, we propose two communication-free parallel MCMC algorithms for sLDA: Simple Average and Weighted Average. In Simple Average, after making predictions for the test dataset based on local topic posteriors for each sub training dataset, we generate the global predictions by taking the arithmetic average of the local predictions. In Weighted Average, after making local predictions in the same way as that in Simple Average, we generate the global predictions by taking the weighted average of the local predictions, where the weights are the inverse of training MSEs (when the labeling variable is continuous) or training accuracy (when the labeling variable is binary. 

We empirically evaluate the performance of our algorithms by comparing our algorithms with non-parallel algorithm and Naive Combination (in which it first combines the sub-posterior and then makes predictions
using the combined global posterior). The empirical results suggest that Simple Average can significantly reduce computation time without sacrificing prediction accuracy.


\bibliographystyle{IEEEtran}
\bibliography{gmproject}

\end{document}